\documentclass{article} 
\input{preamble.tex}
\usepackage{iclr2027_conference,times}

\makeatletter
\def\section{\@startsection {section}{1}{\z@}{-1.6ex plus
    -0.4ex minus -.2ex}{1.1ex plus 0.2ex
minus0.2ex}{\large\sc\raggedright}}
\def\subsection{\@startsection{subsection}{2}{\z@}{-1.4ex plus
-0.4ex minus -.2ex}{0.55ex plus .15ex}{\normalsize\sc\raggedright}}
\def\subsubsection{\@startsection{subsubsection}{3}{\z@}{-1.2ex
plus      -0.4ex minus -.2ex}{0.35ex plus
.15ex}{\normalsize\sc\raggedright}}
\makeatother

\usepackage{hyperref}
\usepackage{url}
\usepackage{amsmath,amssymb,amsfonts,bm,mathtools}
\usepackage{xcolor}
\usepackage{graphicx}
\usepackage{booktabs}
\usepackage{caption}
\usepackage{float}
\usepackage{algorithm}
\usepackage{algpseudocode}
\usepackage{xspace}
\usepackage{multirow}
\usepackage{array}
\usepackage{subcaption}

\newcommand{\ours}{image-structure co-aligned 3D scene generation\xspace}
\newcommand{\oursabbr}{Align3D\xspace}
\newcommand{\oursfull}{\oursabbr: \ours\xspace}
\newcommand{\ourstitle}{\oursabbr: Image-Structure Co-Aligned 3D Scene Generation with Observability-Aware Fidelity-Preserving Guidance}

\newcommand{\sginference}{scene graph inference\xspace}
\newcommand{\sginferencetitle}{Scene Graph Inference}

\newcommand{\coalign}{image-structure co-alignment\xspace}
\newcommand{\coaligntitle}{Image-Structure Co-Alignment}

\newcommand{\ofg}{observability-aware fidelity-preserving guidance\xspace}
\newcommand{\ofgabbr}{OFG\xspace}
\newcommand{\ofgfull}{\ofg\ (\ofgabbr)\xspace}
\newcommand{\ofgtitle}{Observability-Aware Fidelity-Preserving Guidance}

\newcommand{\cwmf}{confidence-weighted modality fusion\xspace}
\newcommand{\cwmfabbr}{CWMF\xspace}
\newcommand{\cwmffull}{\cwmf\ (\cwmfabbr)\xspace}

\title{\ourstitle}

\author{
Chenliang Zhou \\
\And
Yifan Wang \\
\And
Andrzej Banburski-Fahey \\
}

\iclrfinalcopy 

\begin{document}

\maketitle

\vspace{-0.15in}
\begin{figure}[H]
\centering
\includegraphics[width=\textwidth]{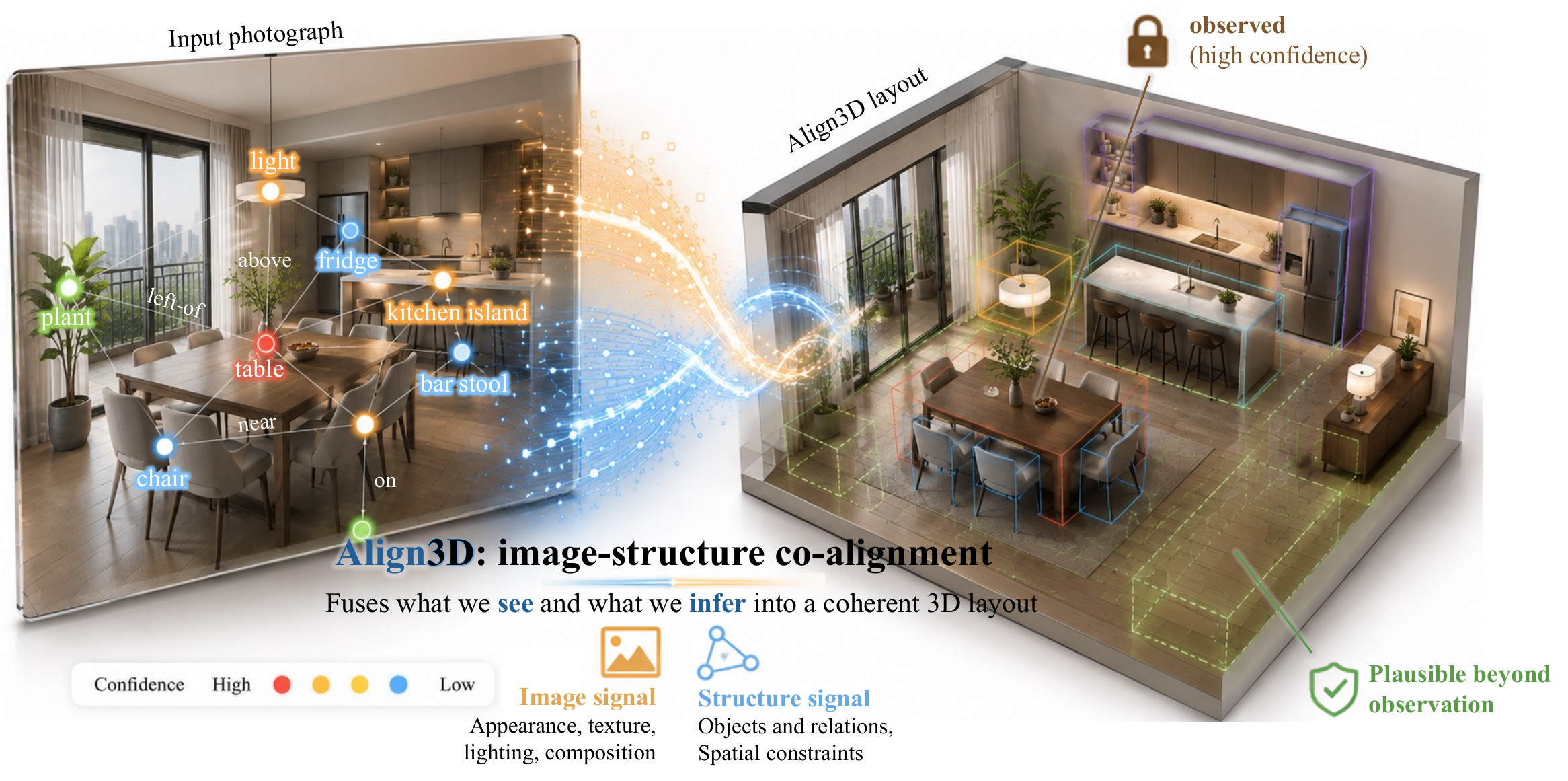}
\caption{\oursabbr couples an image signal and a confidence-calibrated structure signal via \coalign; Observability-aware fidelity-preserving guidance completes under-observed regions under scene-quality guidance while protecting well-supported layout variables, yielding a 
coherent 3D scene.}
\label{fig:teaser}
\end{figure}
\vspace{-0.05in}


\begin{abstract}
We present \oursabbr, an image-conditioned diffusion model that generates 3D indoor layouts that are both faithful to a single input photograph and physically plausible beyond the observed view.
From the photograph we infer a confidence-calibrated scene graph and train a set-diffusion layout denoiser with \coalign where \cwmf balances image and scene-graph alignment using predicted confidences, while \ofg projects conflicting scene-quality gradients and reweights them with an object-level confidence mask so under-observed degrees of freedom absorb physical priors without undoing the image-graph match.
On 3D-FRONT, \oursabbr improves image-alignment and scene-quality metrics over recent baselines.
\end{abstract}


\section{Introduction}
\label{sec:intro}

Generating consistent 3D indoor scenes from a single photograph is a fundamental problem in computer graphics and vision, with applications in embodied AI, virtual staging, and interactive design~\citep{nie2020total3d,sun2018pix3d}.
Early methods recover multi-object geometry via shape retrieval or compositional reconstruction~\citep{han2024reparo,ardelean2025gen3dsr}, while diffusion models now synthesize entire layouts as unordered object sets~\citep{tang2024diffuscene,yang2024physcene}.
Most layout generators condition on text, category priors, or pre-specified scene graphs~\citep{zhai2023commonscenes,wei2024planner3d,feng2026lavigen}; image-to-scene pipelines~\citep{huang2025midi,meng2026scenegen,lin2025partcrafter} ground generation in RGB input but seldom enforce \emph{joint} agreement between visual appearance and inferred spatial structure.

The central challenge is reconciling two complementary signals: \emph{image appearance} (materials, lighting, and composition in the reference view) and \emph{scene structure} (objects and relations that should remain coherent across viewpoints)~\citep{johnson2018image,xu2017scenegraph}.
Visual embeddings anchor layouts to the photograph~\citep{radford2021clip}, and scene graphs encode editable relational constraints~\citep{zhai2023commonscenes,wei2024planner3d}; prior work often optimizes the two sequentially or with fixed loss weights~\citep{ardelean2025gen3dsr,yang2024physcene}, treating every detection and relation as equally reliable even though single-view perception is inherently noisy.
However, neither signal alone guarantees a usable room: a layout can match the image from one camera yet still collide, float, or violate relations that were only weakly observed.

Single-view conditioning sharpens this tension: visible silhouettes pin some objects tightly, while depth ordering, occlusions, and hidden clearances remain underdetermined~\citep{nie2020total3d,sun2018pix3d}.
Physics-aware generators improve plausibility~\citep{yang2024physcene,zhai2023commonscenes,wei2024planner3d}, but typically combine quality terms with conditioning through a weighted sum that can harm image alignment when objectives conflict --- a familiar issue in multi-task learning~\citep{yu2020pcgrad,liu2021cagrad,sener2018mgda}.
Our insight is that observational fidelity and scene quality should not compete along the same parameter directions: strongly observed variables should be protected, while under-constrained ones should absorb physical priors without harming the observation match.
In addition, confidence provides the bookkeeping for this split: down-weighting uncertain detections, camera estimates, and relations during alignment, then routing quality updates away from well-supported layout variables.

To this end, we present \oursfull (\cref{fig:teaser}), an image-conditioned layout diffusion model that predicts a confidence-calibrated scene graph from a photograph and trains with \coalign under \ofg.
Our contributions are:

\begin{itemize}
\item a state-of-the-art image-to-scene pipeline evaluated with competitive baselines on geometry, layout fidelity, and human preference metrics;
\item a confidence-calibrated \sginference module from single-view images that attaches detection, depth, camera, and relational confidences to every graph node and edge, enabling supervision without ground-truth graphs; and
\item end-to-end confidence propagation through \emph{\cwmffull} and \emph{\ofgfull}: \cwmfabbr adaptively balances image-rendering and scene-graph alignment from predicted confidences, while \ofgabbr projects conflicting quality gradients and reweights them with an object-level confidence mask derived from the same predicted confidences.
\end{itemize}


\section{Related Work}
\label{sec:related}

\paragraph{Scene graph representation and learning.}
Scene graphs encode objects and their relations as structured, editable constraints~\citep{johnson2018image,xu2017scenegraph}.
In 3D, early work predicts semantic scene graphs from indoor reconstructions~\citep{wald20203dssg}, while other methods generate and manipulate scenes from given graphs~\citep{dhamo2021graphto3d}.
Subsequent graph-conditioned generators use relational priors for controllable layouts~\citep{zhai2023commonscenes,wei2024planner3d} or couple multimodal graphs with rectified flow~\citep{yang2026flowscene}.
Related text- or instruction-driven layout systems further show that explicit relational structure improves physical plausibility~\citep{feng2026lavigen}, but typically assume language or given objects rather than an observed photograph.
We instead \emph{predict} confidence-calibrated scene graphs from images and use them as \coalign constraints alongside appearance matching.

\paragraph{3D scene generation from images.}
Recovering a multi-object 3D scene from a single RGB image has progressed from feed-forward reconstruction and shape retrieval~\citep{nie2020total3d,sun2018pix3d} to compositional pipelines that segment, complete, and assemble per-object geometry with layout optimization~\citep{ardelean2025gen3dsr,han2024reparo}.
Subsequent systems synthesize multi-object meshes or assets jointly from an image~\citep{huang2025midi,lin2025partcrafter,meng2026scenegen}, while others use depth-guided arrangement~\citep{zhao2025depr}, hierarchical 2D-to-3D room construction~\citep{wang2024architect}, or component-aligned assembly~\citep{yao2025cast}.
Our setting is similar to these baselines, but targets \coalign under object-centric attributes.

\paragraph{3D diffusion models for scenes.}
Denoising diffusion models~\citep{ho2020ddpm,song2021scorebased} generate samples by gradually removing noise from a corrupted signal and have become a leading generative paradigm; transformer backbones further scale this process in latent space~\citep{peebles2023dit}.
Earlier indoor layouts are synthesized autoregressively~\citep{paschalidou2021atiss} or with diffusion over object sets and relational graphs~\citep{tang2024diffuscene,lin2024instructscene,yang2024physcene,bai2025freescene}, while other diffusion models generate room-scale geometry~\citep{ju2024diffindscene}, image-conditioned multi-object meshes~\citep{huang2025midi,lin2025partcrafter}, or vision-guided layouts~\citep{zhu2025imaginarium}.
Most layout-oriented approaches still condition on text, floor plans, or given graphs rather than jointly co-aligning an observed photograph with predicted structure.
We instead train an image-conditioned set-diffusion layout model with \coalign and \ofg, so quality updates improve under-constrained degrees of freedom without undoing the observation match.


\section{Method}
\label{sec:method}

\begin{figure*}[t]
\centering
\includegraphics[width=\textwidth]{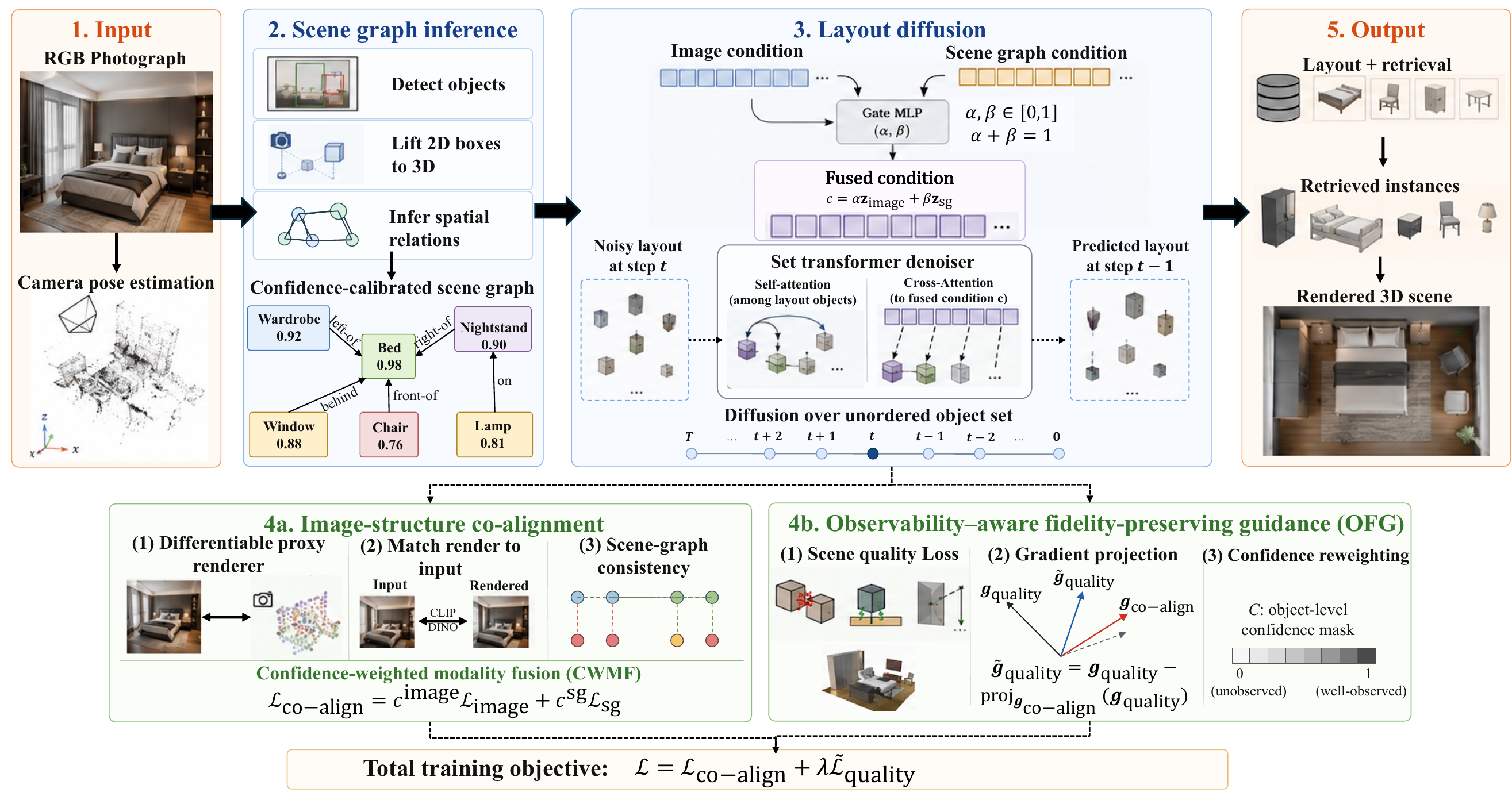}
\caption{\oursabbr pipeline, including
scene graph inference, fused image-graph conditioning, and set-diffusion layout prediction. Training uses \coalign and \ofg, and inference retrieves shapes for rendering.}
\label{fig:pipeline}
\end{figure*}

\oursabbr generates image-conditioned indoor layouts through a set-diffusion denoiser over unordered object configurations (\cref{fig:teaser}).
\Cref{fig:pipeline} summarizes the full system: an input photograph is converted to a confidence-calibrated scene graph, combined with image tokens via fused conditioning, and trained with \coalign and \ofg; after training, the model can map a previously unseen photograph to a layout via the same graph-inference and denoising stack.

Training integrates three main modules: \sginference (\cref{sec:graph_inference}), \coalign (\cref{sec:co_alignment}), and \ofg (\cref{sec:ofg}), which folds a scene-quality loss into optimization under projection and object-level confidence weighting.
The resulting objective is
\begin{equation}
\mathcal{L}
=
\mathcal{L}_{\text{co-align}}
+
\lambda\,
\tilde{\mathcal{L}}_{\text{quality}},
\label{eq:total_loss}
\end{equation}
where $\mathcal{L}_{\text{co-align}}$ couples image and graph observations, $\mathcal{L}_{\text{quality}}$ penalizes inter-object overlap, and $\tilde{\mathcal{L}}_{\text{quality}}$ denotes its fidelity-preserving form with object-level confidence weighting in gradient space.
We detail each module below.
Hyperparameters and implementation details are given in \cref{app:impl}.

\subsection{\sginferencetitle}
\label{sec:graph_inference}

Given a single RGB photograph $I$, we predict a confidence-calibrated scene graph $\mathcal{G}=(\mathcal{V},\mathcal{E})$ that supervises \coalign.
We run open-vocabulary detection with GroundingDINO~\citep{liu2023grounding}: each query names a furniture class from our label vocabulary, and the model returns a category $y_i$, a 2D bounding box $\mathbf{b}_i^{\text{2D}}$, a detection confidence $c_i^{\text{det}}\in[0,1]$, and an appearance embedding $\mathbf{z}_i^{\text{app}}$.
A dense depth map is estimated with Depth Anything~\citep{yang2024depthanything}; we read the estimated depth $d_i$ at the pixel center of each box.
Camera intrinsics $K$, extrinsics $(R,\mathbf{t}^{\text{cam}})$, and a global camera confidence $c^{\text{cam}}\in[0,1]$ are recovered with DUSt3R~\citep{wang2024dust3r}.
We cache the bundle $\mathcal{P}\coloneqq\{K,R,\mathbf{t}^{\text{cam}},c^{\text{cam}}\}$ for 3D lifting and camera-view rendering; $c^{\text{cam}}$ also sets the image-modality weight $c^{\text{image}}$ in image-structure co-alignment (\cref{eq:modality_conf}).

Each detection is lifted to 3D by perspective back-projection.
Let $(u_i,v_i)$ denote the pixel center of $\mathbf{b}_i^{\text{2D}}$, with width $w_i^{\text{2D}}$ and height $h_i^{\text{2D}}$; let $(\bar{u},\bar{v})$ and $(f_x,f_y)$ be the principal point and focal lengths from $K$, and $d_i$ the Depth Anything depth at $(u_i,v_i)$.
We write $\mathbf{p}_i^{\text{cam}}$ for the lifted 3D point in the camera frame and $\mathbf{p}_i^{\text{world}}$ for its center in a shared world frame:
\begin{equation}
\label{eq:lift_3d}
\mathbf{p}_i^{\text{cam}}
\coloneqq
d_i
\begin{bmatrix}
\frac{u_i-\bar{u}}{f_x} \\
\frac{v_i-\bar{v}}{f_y} \\
1
\end{bmatrix},
\qquad
\mathbf{p}_i^{\text{world}}
\coloneqq
R^{\mathrm{T}}(\mathbf{p}_i^{\text{cam}}-\mathbf{t}^{\text{cam}}),
\end{equation}
where $(\cdot)^{\mathrm{T}}$ denotes the transpose.
We estimate coarse 3D extents $\mathbf{w}_i=(w_{i,x},w_{i,y},w_{i,z})^{\top}$ by back-projecting the 2D box ($w_{i,x}=\frac{w_i^{\text{2D}}d_i}{f_x}$, $w_{i,y}=\frac{h_i^{\text{2D}}d_i}{f_y}$) and setting $w_{i,z}\approx\eta\,w_{i,y}$, where $\eta$ is a fixed depth-extent ratio because monocular box height does not constrain object thickness along the viewing axis.
Node confidence aggregates the detection and depth terms with a fixed calibration,
\begin{equation}
c_i^{\text{node}}
\coloneqq
\kappa\, c_i^{\text{det}}\, c_i^{\text{depth}},
\qquad
c_i^{\text{depth}}
\coloneqq
\max\left(0,\,1-\frac{d_i}{d_{\max}}\right),
\end{equation}
where $c_i^{\text{depth}}$ is a depth confidence that decays with distance, $\kappa\in(0,1]$ conservatively reflects residual uncertainty in monocular depth-based 3D lifting, and $d_{\max}$ sets the depth at which $c_i^{\text{depth}}$ reaches zero.
Uncertain detections thus contribute less to later fusion and \ofgabbr.
Spatial relations are inferred from the offset between world-frame centers: letting $\Delta\mathbf{p}_{ij}^{\text{world}} \coloneqq \mathbf{p}_j^{\text{world}}-\mathbf{p}_i^{\text{world}}$, dominant horizontal offsets yield left/right, depth offsets yield front/behind, vertical gaps yield above/on, and nearby contacts yield touching, with proximity radius $r_{\text{close}}=\frac{\|\mathbf{w}_i\|+\|\mathbf{w}_j\|}{2}+\varepsilon_{\text{close}}$.
Edge confidence decays with inter-node distance and is limited by the weaker endpoint,
\begin{equation}
c_{ij}^{\text{edge}}
\coloneqq
\frac{\min(c_i^{\text{node}},c_j^{\text{node}})}{1+\frac{\|\Delta\mathbf{p}_{ij}^{\text{world}}\|}{r_{\text{close}}}}.
\end{equation}
The resulting graph supervises \coalign through node and edge confidences $c_i^{\text{node}}$ and $c_{ij}^{\text{edge}}$, and node-level signals additionally define \ofgabbr's mask $C$ via observability in \cref{sec:ofg}.

\subsection{\coaligntitle}
\label{sec:co_alignment}

\begin{table}[t]
    \centering
    \small
    \setlength{\tabcolsep}{2.2pt}
    \begin{tabular}{lcccccccc}
    \toprule
    Method & CD$\downarrow$ & F-Score$\uparrow$ & IoU$\uparrow$ & CLIP$\uparrow$ & DINO$\uparrow$ & FID$\downarrow$ & HP-Align$\uparrow$ & HP-Scene$\uparrow$ \\
    \midrule
    DiffuScene~\citep{tang2024diffuscene} & 0.292 & 36.1 & 0.289 & 0.784 & 0.758 & 29.8 & 42.8 & 44.1 \\
    Gen3DSR~\citep{ardelean2025gen3dsr} & 0.058 & 59.6 & 0.286 & 0.808 & 0.468 & 52.1 & 46.2 & 47.8 \\
    PartCrafter~\citep{lin2025partcrafter} & 0.194 & 42.0 & 0.236 & 0.814 & 0.552 & 33.8 & 46.8 & 48.1 \\
    MIDI~\citep{huang2025midi} & 0.076 & 52.4 & 0.328 & 0.866 & 0.698 & 23.6 & 51.5 & 50.8 \\
    SceneGen~\citep{meng2026scenegen} & \textbf{0.016} & \underline{87.8} & \underline{0.541} & \underline{0.909} & \underline{0.824} & \textbf{20.3} & \underline{54.2} & \underline{54.8} \\
    \oursabbr (ours) & \underline{0.019} & \textbf{88.7} & \textbf{0.558} & \textbf{0.925} & \textbf{0.843} & \underline{20.8} & \textbf{57.2} & \textbf{56.1} \\
    \bottomrule
    \end{tabular}
    \caption{Main comparison on 3D-FRONT~\citep{fu20213d}.
    \oursabbr is best on image-alignment metrics (CLIP, DINO, HP-Align) and is best or second-best across scene-quality metrics.}
    \label{tab:main}
\end{table}

\coalign trains the denoiser so that generated layouts agree with both the predicted graph from \cref{sec:graph_inference} and the input photograph.
The module first summarizes per-modality reliability, then forms a fused conditioning from image and graph tokens, and finally defines the alignment losses that supervise the predicted layout.

\paragraph{Modality confidences.}
Camera and graph observations can carry unequal reliability across detections.
We aggregate this into scalars
\begin{equation}
c^{\text{image}}
\coloneqq
c^{\text{cam}},
\qquad
c^{\text{sg}}
\coloneqq
\frac{1}{|\mathcal{V}|+|\mathcal{E}|}
\left(
\sum_{i\in\mathcal{V}} c_i^{\text{node}}
+
\sum_{(i,j)\in\mathcal{E}} c_{ij}^{\text{edge}}
\right),
\label{eq:modality_conf}
\end{equation}
where $c^{\text{cam}}$, $c_i^{\text{node}}$, and $c_{ij}^{\text{edge}}$ are defined in \cref{sec:graph_inference}.
Both $c^{\text{image}}$ and $c^{\text{sg}}$ lie in $[0,1]$ and are reused below for both token fusion and loss weighting.

\paragraph{Fused conditioning.}
Following prior set-diffusion layout models~\citep{tang2024diffuscene}, each object is represented by semantic class, oriented bounding box, and a retrievable shape code.
The set-diffusion denoiser is dual-conditioned on the input photograph and the predicted scene graph $\mathcal{G}$.
Let $\mathbf{z}_{\text{img}}$ denote image tokens obtained by projecting CLIP features~\citep{radford2021clip, zhou2023clippae} into the denoiser width and $\mathbf{z}_{\text{sg}}$ denote graph tokens (node and relation embeddings of $\mathcal{G}$).
A small MLP maps the pooled tokens and modality confidences $(c^{\text{image}},c^{\text{sg}})$ to fusion weights $\alpha^{\text{image}},\alpha^{\text{sg}}\in(0,1)$, which form a fused conditioning
\begin{equation}
\mathbf{h}
=
\alpha^{\text{image}}\,\mathbf{z}_{\text{img}}
+
\alpha^{\text{sg}}\,\mathbf{z}_{\text{sg}}.
\label{eq:fused_cond}
\end{equation}
We add a multimodal cross-attention path onto $\mathbf{h}$, so appearance and structure are fused before the alignment losses below are applied.
Because the MLP conditions on $(c^{\text{image}},c^{\text{sg}})$, low graph confidence tends to reduce $\alpha^{\text{sg}}$ and low camera confidence tends to reduce $\alpha^{\text{image}}$.

\paragraph{Alignment losses.}
Generated objects need not be in one-to-one correspondence with predicted graph nodes, so we match them with a soft assignment on node features $\mathbf{f}$ (class embedding concatenated with a linear map of the box).
Letting $\mathbf{f}_i^{\text{obs}}$ and $\mathbf{f}_k^{\text{gen}}$ denote features of observed node $i$ and generated object $k$, the soft matching matrix $M$ is
\begin{equation}
\label{eq:soft_matching}
M_{ik}
=
\mathrm{softmax}_{k}\!\left(\frac{\cos(\mathbf{f}_i^{\text{obs}},\mathbf{f}_k^{\text{gen}})}{\tau}\right),
\end{equation}
where $\mathrm{softmax}_{k}$ is the softmax over generated index $k$, $\tau>0$ is a temperature that sharpens the assignment, and $M_{ik}$ is the probability of matching $i$ to $k$.
Let $\mathbf{t}_k$ and $\mathbf{s}_k$ denote the translation and size of generated object $k$.
The scene-graph loss then averages differentiable soft relation predicates over observed edges $(i,r,j)\in\mathcal{E}$, where $r$ denotes the relation type (left, right, front, behind, above, \etc).
Transporting each predicate onto matched generated pairs gives
\begin{equation}
\mathcal{L}_{\text{sg}}
=
\frac{1}{|\mathcal{E}|}
\sum_{(i,r,j)\in\mathcal{E}}
\sum_{k,m}
M_{ik}\,M_{jm}\,
\ell_r(\mathbf{t}_k,\mathbf{t}_m,\mathbf{s}_k,\mathbf{s}_m),
\end{equation}
where $\ell_r$ is a soft hinge for relation $r$ (definitions in \cref{app:impl}).
Here $M$ lets gradients flow through uncertain matches rather than forcing a hard permutation.

For appearance we do not back-propagate through the full mesh rasterizer.
To keep training efficient we render the current layout as a differentiable 2D Gaussian-splat proxy image $I_{\text{proxy}}$ by projecting each generated center into the image under the camera bundle $\mathcal{P}$.
Writing $\mathbf{p}_k^{\text{cam}}\coloneqq R\mathbf{t}_k+\mathbf{t}^{\text{cam}}$ and $z_k\coloneqq(\mathbf{p}_k^{\text{cam}})_z$, the projected center and image-plane scales are
\begin{equation}
\mathbf{u}_k
=
\begin{bmatrix}
\bar{u}+f_x\,(\mathbf{p}_k^{\text{cam}})_x/z_k \\
\bar{v}+f_y\,(\mathbf{p}_k^{\text{cam}})_y/z_k
\end{bmatrix},
\qquad
\boldsymbol{\sigma}_k
=
\begin{bmatrix}
f_x\,(\mathbf{s}_k)_x/z_k \\
f_y\,(\mathbf{s}_k)_y/z_k
\end{bmatrix},
\end{equation}
and the proxy has pixel values
\begin{equation}
I_{\text{proxy}}(u,v)
=
\sum_k
\mathbf{a}_k
\exp\left(
-\frac{\left(u-(\mathbf{u}_k)_x\right)^2}{2\left((\boldsymbol{\sigma}_k)_x\right)^2}
-\frac{\left(v-(\mathbf{u}_k)_y\right)^2}{2\left((\boldsymbol{\sigma}_k)_y\right)^2}
\right),
\end{equation}
where $\mathbf{a}_k$ is a learnable class color (RGB).
Let $E_{\text{CLIP}}$ and $E_{\text{DINO}}$ denote the image encoders of CLIP~\citep{radford2021clip} and DINOv2~\citep{oquab2024dinov2}.
The image alignment loss matches the proxy to the input photograph in both embedding spaces,
\begin{equation}
\mathcal{L}_{\text{image}}
=
1-\cos\left(E_{\text{CLIP}}(I_{\text{proxy}}),E_{\text{CLIP}}(I)\right)
+
\lambda_{\text{dino}}
\left(
1-\cos\left(E_{\text{DINO}}(I_{\text{proxy}}),E_{\text{DINO}}(I)\right)
\right),
\end{equation}
where $\lambda_{\text{dino}}>0$ balances the two embedding terms.

Finally, we propose the \emph{\cwmffull} to combine the two observational terms with the modality confidences from \cref{eq:modality_conf}:
\begin{equation}
\mathcal{L}_{\text{co-align}}
=
c^{\text{image}}\,\mathcal{L}_{\text{image}}
+
c^{\text{sg}}\,\mathcal{L}_{\text{sg}}.
\label{eq:cwmf}
\end{equation}
When one modality is unreliable, \cwmfabbr automatically down-weights its loss relative to the other.
  
\begin{figure*}[t]
    \centering
    \includegraphics[width=\textwidth]{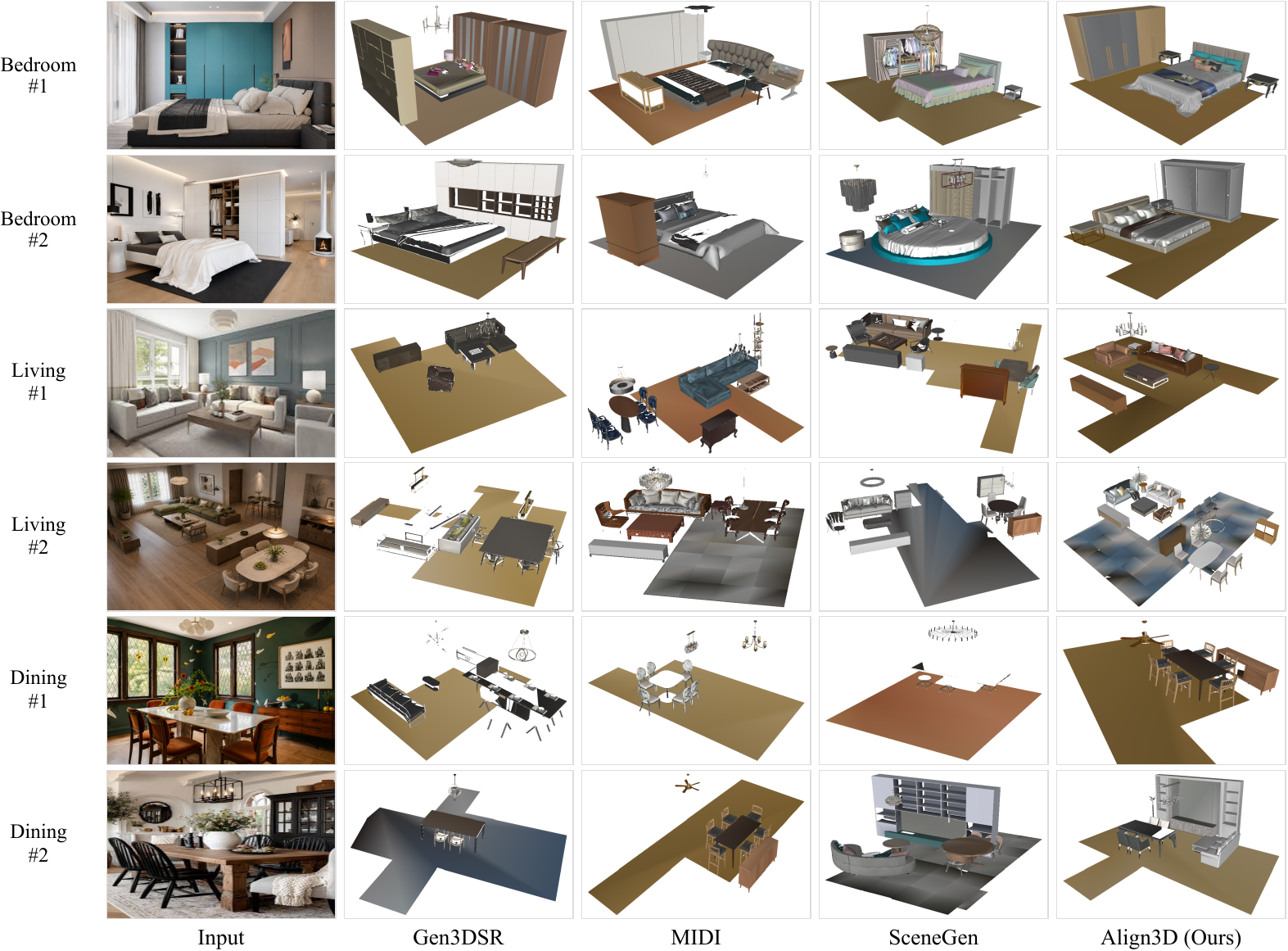}
    \caption{Qualitative image-to-scene comparison on real-world indoor scenes.
    Each row shows the input photograph and matched-view renderings from different methods.
    \oursabbr better matches the observed furniture arrangement and reduces collisions and floating objects relative to the baselines.}
    \label{fig:qualitative_main}
\end{figure*}

\subsection{\ofgtitle}
\label{sec:ofg}

Image-structure co-alignment establishes observational fidelity but does not by itself resolve residual physical ambiguity outside the observed silhouette.
Therefore, inspired by DiffuScene~\citep{tang2024diffuscene}, we introduce a scene-quality loss that penalizes pairwise oriented-box interpenetration,
\begin{equation}
\mathcal{L}_{\text{quality}}
=
\sum_{i<j}
\operatorname{IoU}(i,j),
\label{eq:quality_loss}
\end{equation}
where $\operatorname{IoU}(i,j)$ is the volumetric intersection-over-union of the oriented boxes of objects $i$ and $j$~\citep{paschalidou2021atiss}.

Nevertheless, simply adding $\mathcal{L}_{\text{quality}}$ to $\mathcal{L}_{\text{co-align}}$ can undo the image-graph match: when the two gradients conflict, the quality term pulls the layout toward a generic collision-free arrangement rather than the observed scene.
We therefore introduce \emph{\ofgfull}, which folds $\mathcal{L}_{\text{quality}}$ into training under fidelity-preserving, confidence-aware updates.

Let $\mathbf{g}_{\text{co-align}}\coloneqq\nabla\mathcal{L}_{\text{co-align}}$ and $\mathbf{g}_{\text{quality}}\coloneqq\nabla\mathcal{L}_{\text{quality}}$.
Adapting gradient projection from multi-task learning~\citep{yu2020pcgrad,liu2021cagrad,sener2018mgda}, we treat \coalign as primary and remove the conflicting component of the quality gradient:
\begin{equation}
\tilde{\mathbf{g}}_{\text{quality}}
=
\begin{cases}
\mathbf{g}_{\text{quality}}-\dfrac{\mathbf{g}_{\text{quality}}^{\mathrm{T}} \mathbf{g}_{\text{co-align}}}{\|\mathbf{g}_{\text{co-align}}\|^2+\epsilon}\,\mathbf{g}_{\text{co-align}},
&
\mathbf{g}_{\text{quality}}^{\mathrm{T}} \mathbf{g}_{\text{co-align}} < -\xi,\\[8pt]
\mathbf{g}_{\text{quality}},
&
\text{otherwise,}
\end{cases}
\label{eq:project_quality}
\end{equation}
and write $\tilde{\mathcal{L}}_{\text{quality}}$ for the quality term under this projected gradient.
Here $\xi\ge 0$ is a conflict margin and $\epsilon>0$ is a numerical stabilizer.
By construction, under conflict $\mathbf{g}_{\text{co-align}}^{\mathrm{T}}\tilde{\mathbf{g}}_{\text{quality}}\approx 0$, so a step along $\mathbf{g}_{\text{co-align}}+\lambda\tilde{\mathbf{g}}_{\text{quality}}$ does not increase $\mathcal{L}_{\text{co-align}}$ to first order, while quality still improves in directions orthogonal to alignment.

Moreover, a single view constrains objects unequally: well-observed silhouettes are tightly pinned, while occluded depth and hidden clearances remain free.
We summarize this for each observed graph node by an \emph{observability} score $o_i\in[0,1]$,
\begin{equation}
o_i
\coloneqq
\mathrm{sigmoid}\left(
\alpha^{\text{vis}}\left(2\nu_i-1\right)
+
\alpha^{\text{node}}\left(2c_i^{\text{node}}-1\right)
\right),
\label{eq:obj_conf}
\end{equation}
where $\mathrm{sigmoid}(x)\coloneqq(1+e^{-x})^{-1}$, $\nu_i\in[0,1]$ is a visibility score (details in \cref{app:impl}), $c_i^{\text{node}}$ is the node confidence from \cref{sec:graph_inference}, and $\alpha^{\text{vis}}$, $\alpha^{\text{node}}$ are fixed observability coefficients.
We then transport observability through the matching matrix $M$ (\cref{eq:soft_matching}) to obtain the observability score for each generated object:
\begin{equation}
o_k^{\text{gen}}
\coloneqq
\sum_i M_{ik}\,o_i,
\end{equation}
and partition layout parameters into $N$ generated-object blocks with $C\coloneqq\operatorname{diag}(1-o_1^{\text{gen}},\ldots,1-o_N^{\text{gen}})$ to scale each block of the projected quality gradient, concentrating quality updates on less observable objects:
\begin{equation}
\mathbf{g}
=
\mathbf{g}_{\text{co-align}}
+
\lambda\,
C\,\tilde{\mathbf{g}}_{\text{quality}},
\label{eq:combined_grad}
\end{equation}
which implements \cref{eq:total_loss} with object-level confidence weighting in gradient space.
Intuitively, generation freedom is proportional to residual unobservability: high-$o_k^{\text{gen}}$ objects should barely move under quality updates, while low-$o_k^{\text{gen}}$ ones should absorb the physical prior.

\begin{table}[t]
    \centering
    \small
    \setlength{\tabcolsep}{2.2pt}
    \begin{tabular}{@{}lcccccc@{}}
    \toprule
    Variant & CD$\downarrow$ & F-Score$\uparrow$ & IoU$\uparrow$ & CLIP$\uparrow$ & DINO$\uparrow$ & FID$\downarrow$ \\
    \midrule
    Quality loss: $\lambda\mathcal{L}_{\text{quality}}$ & 0.286 & 36.8 & 0.312 & 0.786 & 0.754 & 29.1 \\
    Image alignment: $\mathcal{L}_{\text{image}}+\lambda\mathcal{L}_{\text{quality}}$ & 0.068 & 76.4 & 0.438 & 0.896 & 0.816 & 23.9 \\
    SG alignment: $\mathcal{L}_{\text{sg}}+\lambda\mathcal{L}_{\text{quality}}$ & 0.044 & 81.2 & 0.516 & 0.838 & 0.782 & 25.4 \\
    Co-alignment: $\mathcal{L}_{\text{image}}+\mathcal{L}_{\text{sg}}+\lambda\mathcal{L}_{\text{quality}}$ & \underline{0.027} & \underline{86.4} & \underline{0.542} & \underline{0.898} & \underline{0.818} & \underline{22.1} \\
    \cwmfabbr: $\mathcal{L}_{\text{co-align}}+\lambda\mathcal{L}_{\text{quality}}$ & \textbf{0.025} & \textbf{86.8} & \textbf{0.554} & \textbf{0.901} & \textbf{0.822} & \textbf{21.2} \\
    \bottomrule
    \end{tabular}
    \caption{Image-structure co-alignment ablation.
    Quality loss alone cannot ground layouts in the input; image and scene-graph terms are each necessary for complementary visual and structural gains, and confidence-weighted \cwmfabbr introduces further improvements over fixed-weight fusion.}
    \label{tab:ablation_coalign}
\end{table}


\section{Experiments}
\label{sec:experiments}

\subsection{Dataset and Baselines}

We evaluate on 3D-FRONT~\citep{fu20213d}, a large-scale dataset of synthetic indoor scenes with object layouts and room geometry
Following ATISS~\citep{paschalidou2021atiss} and DiffuScene~\citep{tang2024diffuscene}, we use three room types---4{,}041 bedrooms, 900 dining rooms, and 813 living rooms---with an 80/20 train/test split per type.
Conditioning inputs are single rendered RGB views of each room; for the quantitative protocol we use the corresponding rendering cameras when available.

We compare \oursabbr against five recent baselines under a unified protocol: DiffuScene~\citep{tang2024diffuscene}, Gen3DSR~\citep{ardelean2025gen3dsr}, PartCrafter~\citep{lin2025partcrafter}, MIDI~\citep{huang2025midi}, and SceneGen~\citep{meng2026scenegen}.
For DiffuScene, we use an image-conditioned adaptation of the official denoiser for fair single-image comparison (details in \cref{app:impl}).
Baselines use official code and default settings when available; all methods are evaluated on the same conditioning viewpoint.

\subsection{Metrics}

We report two families of automatic metrics: scene quality and image alignment.
Scene-quality metrics are CD$\downarrow$ (Chamfer distance between point sets sampled from generated and ground-truth scene surfaces)~\citep{fan2017point}, F-Score$\uparrow$ (harmonic mean of precision and recall at a fixed Euclidean distance threshold)~\citep{tatarchenko2019what}, IoU$\uparrow$ (matched object-box intersection-over-union \vs ground truth)~\citep{paschalidou2021atiss}, and FID$\downarrow$ (Fr\'echet distance between inception features of matched-view generated renderings and ground-truth test-set renderings)~\citep{heusel2017fid}.
Image-alignment metrics are CLIP$\uparrow$ and DINO$\uparrow$ (cosine similarity between global CLIP~\citep{radford2021clip} and DINOv2~\citep{oquab2024dinov2} embeddings of the conditioning image and a matched-view rendering).
In addition, we introduce survey-based metrics, defined next.
Unless noted otherwise, the best result in each table is set in \textbf{bold} and the second best is \underline{underlined}.

\paragraph{Human preference study.}
Automatic metrics do not fully capture perceived image match or placement plausibility, so we run a pairwise preference study.
Each trial shows the conditioning image and two anonymized matched-view renderings from two methods drawn uniformly at random from DiffuScene, Gen3DSR, PartCrafter, MIDI, SceneGen, and our \oursabbr; the participant answers two questions: (i)~which layout is closer to the input image, and (ii)~which placement is more physically reasonable.
We recruit 40 participants; each answers 20 trials (40 binary choices), yielding 800 comparisons per criterion.
Order of methods and left/right placement are randomized, and ties are disallowed.
We then fit a Bradley-Terry model~\citep{bradley1952rank} independently for each criterion: each method $m$ receives a latent strength $\pi_m$, and the model predicts that $m$ is chosen over $m'$ with probability $P(m \succ m')=\frac{\pi_m}{\pi_m+\pi_{m'}}$.
After maximum-likelihood fitting on all pairwise choices, we convert each $\pi_m$ to a report score on a 0--100 scale where the cohort average maps to $50$.
We denote the resulting scores for questions~(i) and~(ii) by HP-Align$\uparrow$ (human preference for image alignment) and HP-Scene$\uparrow$ (human preference for scene quality), respectively.
These scores indicate how often participants preferred a method relative to the pool average (\eg, $57$ means modestly above-average preference).

\begin{table}[t]
    \centering
    \small
    \setlength{\tabcolsep}{2.0pt}
    \begin{tabular}{@{}lcccccc@{}}
    \toprule
    Variant & CD$\downarrow$ & F-Score$\uparrow$ & IoU$\uparrow$ & CLIP$\uparrow$ & DINO$\uparrow$ & FID$\downarrow$ \\
    \midrule
    Sum: $\mathbf{g}_{\text{co-align}}+\lambda\mathbf{g}_{\text{quality}}$ & 0.025 & 86.8 & 0.554 & 0.901 & 0.822 & 21.2 \\
    Projection: $\mathbf{g}_{\text{co-align}}+\lambda\tilde{\mathbf{g}}_{\text{quality}}$ & \underline{0.020} & \underline{88.1} & \underline{0.555} & \underline{0.921} & \underline{0.840} & 21.0 \\
    Reversed projection: $\tilde{\mathbf{g}}_{\text{co-align}}+\lambda\mathbf{g}_{\text{quality}}$ & 0.031 & 84.6 & \textbf{0.571} & 0.866 & 0.798 & \textbf{20.5} \\
    \ofgabbr: $\mathbf{g}_{\text{co-align}}+\lambda\,C\tilde{\mathbf{g}}_{\text{quality}}$ & \textbf{0.019} & \textbf{88.7} & \underline{0.558} & \textbf{0.925} & \textbf{0.843} & \underline{20.8} \\
    \bottomrule
    \end{tabular}
    \caption{\ofgabbr ablation.
    Unconstrained summation hurts alignment; projection is necessary to avoid gradient conflict, and confidence mask $C$ introduces further improvements.}
    \label{tab:ablation_ofpg}
\end{table}

\subsection{Results}

\Cref{tab:main} reports the main quantitative comparison.
\oursabbr is best on image-alignment metrics (CLIP, DINO, HP-Align) and is best or second-best across scene-quality metrics.
\Cref{fig:qualitative_main} shows six real-world indoor scenes (two per room type) on held-out photographs (more in \cref{app:qualitative}).
Matched-view renderings use 3D-FUTURE~\citep{fu20213dfuture} assets, so we advise comparing layout and placement rather than surface color.
Across room types, \oursabbr better recovers the observed arrangement with fewer artifacts such as collisions and floating furniture than the baselines.

\subsection{Ablation Studies}
\label{sec:ablation}


\paragraph{Co-alignment components.}
We study how different alignment terms interact with the scene-quality loss and the impact of confidence weighting. In \cref{tab:ablation_coalign}, each row lists the variant name followed by the training objective it uses.
Training with $\mathcal{L}_{\text{quality}}$ alone (``Quality loss'') yields weak image-scene agreement (CLIP 0.786, DINO 0.754), confirming that a generic layout prior is insufficient for image-conditioned generation.
Adding image alignment (``Image alignment'') is essential for visual fidelity: CLIP and DINO increase to 0.896 and 0.816 respectively, but IoU remains low (0.438) because appearance alone does not constrain relational layout.
Conversely, adding scene-graph alignment (``SG alignment'') is essential for structure: IoU rises to 0.516 and CD drops to 0.044, while CLIP lags behind the image-only variant.
Combining both terms (``Co-alignment'') is required to capture both effects, yet fixed equal weighting still underperforms \cwmfabbr on every metric, showing that propagating predicted confidences into the alignment objective is necessary when graphs or camera estimates are noisy.

\paragraph{\ofgabbr components.}
We now investigate different ways to combine $\mathcal{L}_{\text{quality}}$ with $\mathcal{L}_{\text{co-align}}$.
In \cref{tab:ablation_ofpg}, each row lists the variant name followed by the gradient update it uses: ``Sum'' adds unconstrained quality gradients; ``Projection'' replaces them by the fidelity-preserving $\tilde{\mathbf{g}}_{\text{quality}}$; ``Reversed projection'' instead protects quality and projects \coalign away from it; ``\ofgabbr'' applies the confidence mask $C$ to the projected quality gradient.
Naively summing both losses (``Sum'') degrades image alignment relative to projection (CLIP 0.901 \vs 0.921), indicating that gradient projection is necessary whenever quality and \coalign objectives conflict.
Reversed projection confirms the design choice: protecting $\mathcal{L}_{\text{quality}}$ rather than co-alignment yields higher IoU (0.571) at the cost of substantially worse alignment metrics, \ie sacrificing observation fidelity for generic plausibility.
Full \ofgabbr is necessary for the best overall tradeoff---it matches or exceeds projection on alignment metrics while delivering the strongest CD and F-Score, because $C$ routes quality gradients to under-observed objects instead of overwriting image-graph evidence on well-constrained instances.

\subsection{Efficiency Analysis}
\label{sec:efficiency}

\begin{figure}[t]
\centering
\includegraphics[width=\linewidth]{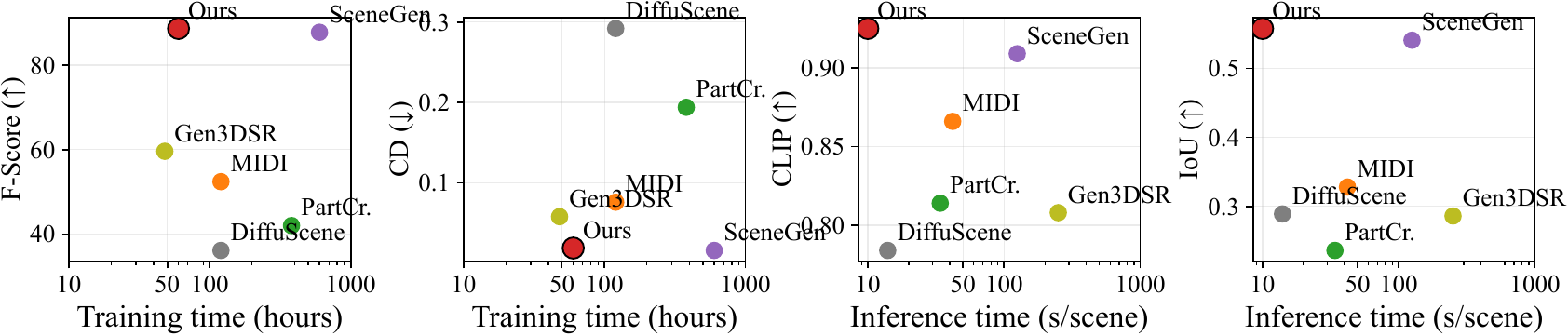}
\caption{Quality \vs efficiency plots.
Our method achieves both high quality and fast runtime.}
\label{fig:efficiency_scatter}
\end{figure}

Beyond accuracy, practical systems must also be efficient.
In \cref{tab:efficiency}, we compare training and per-scene inference times on a single NVIDIA RTX~4090.
At inference, \oursabbr runs graph prediction (${\sim}3$\,s), 50-step layout diffusion (${\sim}5$\,s), and scene composition (${\sim}2$\,s), totaling ${\sim}10$\,s per scene, the fastest among compared methods. \Cref{fig:efficiency_scatter} also plots the metrics against training or inference time.
\oursabbr occupies the fast regime on both axes while remaining competitive or best on scene-quality and image-alignment metrics, illustrating a favorable quality-efficiency tradeoff for layout-centric image-to-scene generation.

\begin{table}[t]
\centering
\small
\setlength{\tabcolsep}{4pt}
\begin{tabular}{@{}lcc@{}}
\toprule
Method & Training (h) & Inference (s/scene) \\
\midrule
DiffuScene~\citep{tang2024diffuscene} & 120 & \underline{14} \\
Gen3DSR~\citep{ardelean2025gen3dsr} & \textbf{48} & 250 \\
PartCrafter~\citep{lin2025partcrafter} & 380 & 34 \\
MIDI~\citep{huang2025midi} & 120 & 42 \\
SceneGen~\citep{meng2026scenegen} & 600 & 125 \\
\oursabbr (ours) & \underline{60} & \textbf{10} \\
\bottomrule
\end{tabular}
\caption{Training and inference cost. Our method is fast while producing high quality results.}
\label{tab:efficiency}
\end{table}

\section{Conclusion}

We presented \oursabbr, which trains an image-conditioned layout generator with \coalign and \ofgabbr so that generated rooms both explain the observation and remain physically plausible.
The training objective protects image-graph evidence while letting scene-quality updates act in under-observed degrees of freedom.
Limitations remain: we evaluate only single-image conditioning (\cref{app:multiview} sketches a preliminary multi-view extension), proxy rendering does not take colors into account, and errors in predicted scene graphs can propagate into \coalign. We leave these limitations for future work.


\clearpage

\section*{AI use statement}
In this work, we used large language models to polish the writing of the paper and to help debug the implementation.
We have reviewed all AI-assisted text and code changes and take full responsibility for the final content.

\section*{Ethics statement}
This work studies indoor scene generation on publicly available datasets.
The human preference study collects anonymous pairwise judgments about generated layouts and does not collect personally identifying information.
Generated layouts are intended for research on graphics and embodied-AI simulation; we discourage misuse for fabricating misleading real-estate or surveillance content.

\section*{Reproducibility statement}
Code is provided in the supplementary materials.
Implementation details and hyperparameters are given in \cref{app:impl}.
All quantitative protocols and ablations are described in \cref{sec:experiments}.


\bibliographystyle{iclr2027_conference}
\bibliography{refs}


\appendix
\clearpage

\section{Implementation Details}
\label{app:impl}

\paragraph{Layout denoiser.}
We instantiate the set-diffusion layout denoiser following DiffuScene~\citep{tang2024diffuscene}: an image-conditioned denoiser over unordered object tuples (semantics, oriented boxes, shape codes) with shape retrieval from a shape library (3D-FUTURE~\citep{fu20213dfuture} in our experiments).
Training uses Adam with learning rate $10^{-3}$, batch size $128$, and $60{,}000$ iterations; the diffusion process uses a linear $\beta$ schedule from $10^{-4}$ to $2{\times}10^{-2}$ over $T{=}1000$ steps, with $50$-step DDIM sampling at inference.
For the DiffuScene baseline, the official unconditional layout denoiser is adapted by encoding the conditioning RGB with CLIP~\citep{radford2021clip} and injecting the resulting image embedding as an additional condition into the U-Net (cross-attention over the layout tokens), then training on the same 3D-FRONT splits with matched rendered views while keeping the original network width, diffusion schedule, and retrieval head otherwise unchanged.

\paragraph{Scene graph prediction.}
Object detection uses GroundingDINO~\citep{liu2023grounding} with the 23-class vocabulary from 3D-FRONT.
Depth estimation uses Depth Anything~\citep{yang2024depthanything} (base), and camera pose is estimated with DUSt3R~\citep{wang2024dust3r}.
For the 3D-FRONT evaluation protocol (\cref{sec:experiments}), conditioning images are rendered views and we optionally substitute known rendering cameras for $K$ and $(R,\mathbf{t}^{\text{cam}})$ with $c^{\text{cam}}{=}1$.
Scene-graph hyperparameters in \cref{sec:graph_inference} are $\kappa{=}0.8$, $\eta{=}0.5$, $d_{\max}{=}5$\,m, and $\varepsilon_{\text{close}}{=}0.2$\,m.
At export, when the generated layout and inferred graph disagree in global yaw, we optionally apply a closed-form floor-plane rotation to align layout translations to graph centers before camera-view rendering; individual layout boxes are not edited.

\paragraph{Training losses.}
Unless otherwise stated, $\lambda_{\text{dino}}{=}0.3$ and the soft-matching temperature $\tau{=}0.07$.
\ofgabbr uses $\lambda{=}0.15$ on $\mathcal{L}_{\text{quality}}$, project whenever $\mathbf{g}_{\text{co-align}}^{\mathrm{T}}\mathbf{g}_{\text{quality}}<0$ ($\xi{=}0$, $\epsilon{=}10^{-8}$), and set $\alpha^{\text{vis}}{=}1.0$, $\alpha^{\text{node}}{=}0.5$ in \cref{eq:obj_conf}.
Projection always protects \coalign except in the reverse-projection ablation.
For scene-quality metrics in \cref{sec:experiments}, F-Score uses a Euclidean distance threshold of $0.05$\,m.

\paragraph{Relation hinges.}
Writing $(\mathbf{t}_k)_x$ for the $x$-component of generated translation $\mathbf{t}_k$, the left hinge is
$\ell_{\text{left}}(\mathbf{t}_k,\mathbf{t}_m)\coloneqq\mathrm{softplus}\!\left((\mathbf{t}_k)_x-(\mathbf{t}_m)_x+\delta\right)$
with $\mathrm{softplus}(x)\coloneqq\log(1+e^{x})$ and margin $\delta{=}0.05$; minimizing it encourages object $k$ to lie left of $m$.
Analogous softplus hinges cover right, front, behind, and above.

\paragraph{Visibility.}
Visibility $\nu_i\in[0,1]$ in \cref{eq:obj_conf} is the fraction of object $i$'s 2D box area not occluded by other objects closer to the camera after sorting by depth $d_i$; when depth ordering is unavailable we set $\nu_i{=}1$.

\section{Additional Qualitative Results}
\label{app:qualitative}

\begin{figure*}[t]
\centering
\includegraphics[width=\textwidth]{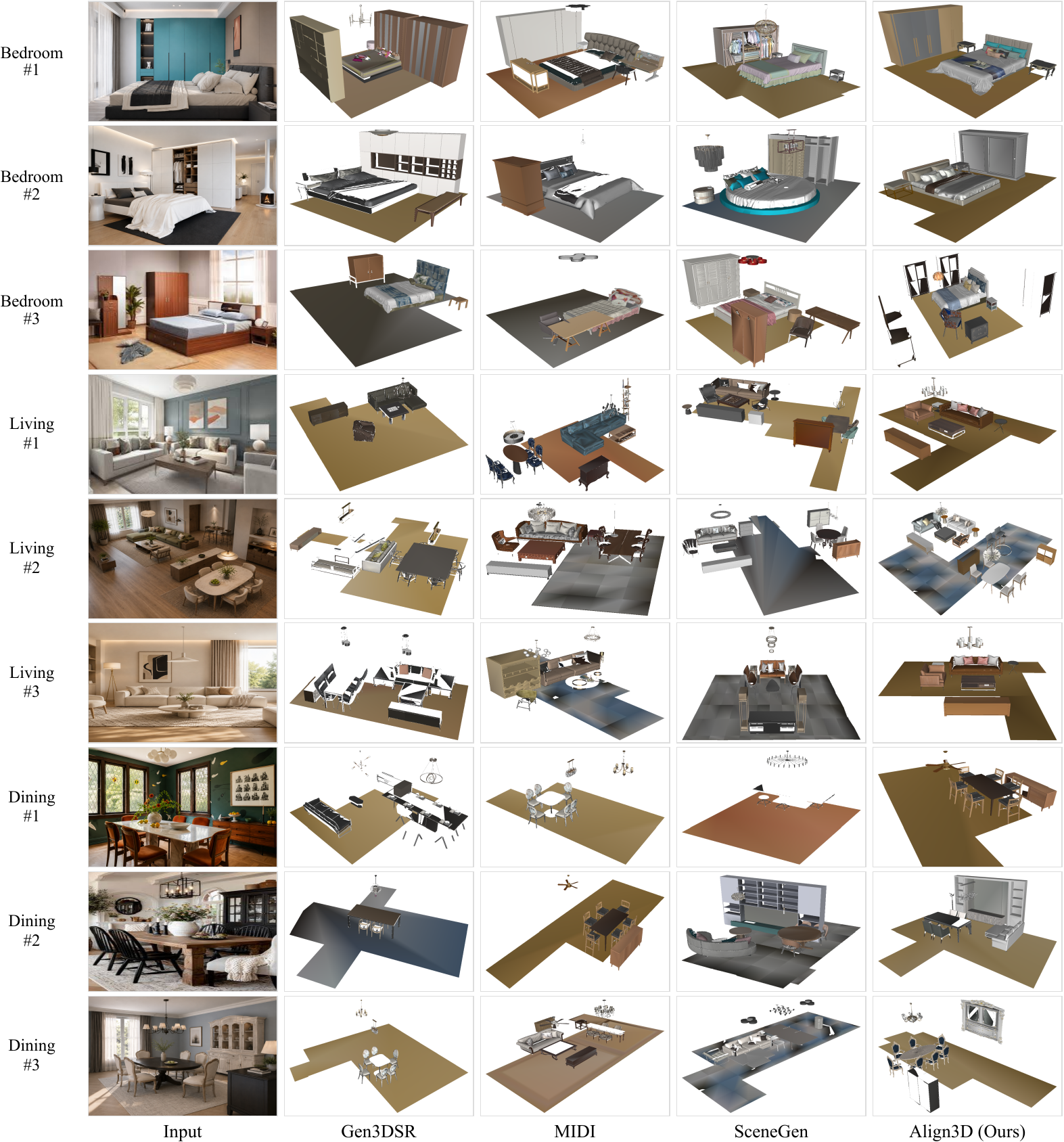}
\caption{Qualitative comparison on nine real-world indoor scenes.}
\label{fig:qualitative_all}
\end{figure*}

\Cref{fig:qualitative_all} extends \cref{fig:qualitative_main} to nine real-world evaluation scenes (three per room type).

\section{Multi-View Extension}
\label{app:multiview}

All experiments in the main paper condition on a \emph{single} RGB photograph.
This section documents a multi-view generalization that reuses the same confidence-calibrated design: additional views tighten graph inference and image alignment where geometry is ambiguous in any one frame, while unreliable views are down-weighted rather than averaged uniformly.
We include it to make the single-view presentation in \cref{sec:method} explicit about scope and to record the extension path.

\paragraph{Scene graph inference.}
Given $V$ RGB views $\{I_v\}_{v=1}^{V}$, we run GroundingDINO independently on each view and estimate cameras with DUSt3R~\citep{wang2024dust3r}, yielding per-view intrinsics $K_v$, extrinsics $(R_v,\mathbf{t}_v^{\text{cam}})$, and camera confidence $c_v^{\text{cam}}\in[0,1]$.
Each 2D detection is lifted to world coordinates with the same back-projection as \cref{sec:graph_inference}, using $(K_v,R_v,\mathbf{t}_v^{\text{cam}})$ and Depth Anything depth from the view.
Node confidence in view $v$ follows the same $\kappa$-weighted detection--depth product as in \cref{sec:graph_inference}.

Cross-view instances of the same physical object must be merged before relation inference.
We greedily group same-class detections from different views when their combined similarity exceeds a threshold $\tau_{\text{match}}{=}0.5$:
\begin{equation}
s_{ij}
\coloneqq
\lambda_{\text{app}}\,\cos(\mathbf{z}_i^{\text{app}},\mathbf{z}_j^{\text{app}})
+
\lambda_{\text{geo}}\,\frac{1}{1+\frac{d_{ij}^{\text{3D}}}{r_{\text{merge}}}},
\end{equation}
where $d_{ij}^{\text{3D}}\coloneqq\|\mathbf{p}_i^{\text{world}}-\mathbf{p}_j^{\text{world}}\|$, $r_{\text{merge}}{=}0.5$\,m normalizes the proximity term, and $\lambda_{\text{app}}{=}0.6$, $\lambda_{\text{geo}}{=}0.4$ with $\lambda_{\text{app}}+\lambda_{\text{geo}}{=}1$ balance appearance and geometry.
Matched detections are fused by confidence-weighted averaging of 3D position and size; fused node confidence is the maximum view confidence (scaled conservatively) so that agreement across views increases structural trust.
Spatial relations and edge confidences are then inferred on the fused graph exactly as in \cref{sec:graph_inference}.

\paragraph{Image-structure co-alignment and fused conditioning.}
Modality confidences generalize to
\begin{equation}
c^{\text{image}}
\coloneqq
\max_{v\in\{1,\ldots,V\}} c_v^{\text{cam}},
\end{equation}
so a reliable camera estimate preserves image-modality trust even if other views are uncertain.
Image tokens $\mathbf{z}_{\text{img}}$ are extracted per view (CLIP projection as in \cref{sec:co_alignment}) and mean-pooled before the fused conditioning in \cref{eq:fused_cond}.
For appearance alignment, we render the layout proxy from each estimated camera and average view losses with confidence weights:
\begin{equation}
\mathcal{L}_{\text{image}}
=
\frac{\sum_{v=1}^{V}c_v^{\text{cam}}\,\mathcal{L}_{\text{image}}^{(v)}}{\sum_{v=1}^{V}c_v^{\text{cam}}},
\end{equation}
where $\mathcal{L}_{\text{image}}^{(v)}$ is the per-view CLIP/DINO alignment term defined in \cref{sec:co_alignment} for reference image $I_v$.
When $V{=}1$, these expressions reduce to the single-image formulation in \cref{sec:co_alignment}.
Scene-graph alignment, \cwmf, and \ofgabbr are unchanged aside from the richer graph and image terms above.

\paragraph{Inference-time multi-view guidance.}
At sampling time we additionally support a multi-view Gaussian-splat renderer that projects the current layout into each estimated camera before computing CLIP/DINO guidance, mirroring the training-time proxy but with per-view $K_v,R_v,\mathbf{t}_v^{\text{cam}}$.
This path is useful for in-the-wild inputs where DUSt3R provides the camera bundle $\mathcal{P}$ and multiple photographs of the same room are available.
Confidence-weighted fusion of per-view guidance gradients follows the same $c_v^{\text{cam}}$ weighting as $\mathcal{L}_{\text{image}}$.

\end{document}